\documentclass[letterpaper]{article} 
\usepackage{aaai2026}  
\usepackage{times}  
\usepackage{helvet}  
\usepackage{courier}  
\usepackage[hyphens]{url}  
\usepackage{graphicx} 
\usepackage{natbib}  
\usepackage{caption} 
\usepackage{algorithm}
\usepackage{algorithmic}

\usepackage{amssymb} 
\usepackage{xcolor} 
\usepackage[table,xcdraw]{xcolor} 
\usepackage{booktabs} 
\usepackage{array} 
\usepackage{subcaption} 
\usepackage{amsmath}

\usepackage{newfloat}
\usepackage{listings}

\definecolor{myhighlightcolor}{HTML}{F7E7CE}

\makeatletter
\newcommand{\ColorState}[2]{%
  \STATE\makebox[0pt][l]{%
    \hspace{-\algorithmicindent}%
    \color{#1}%
    \rule[-3pt]{\dimexpr\linewidth+\algorithmicindent\relax}{1.1\baselineskip}%
  }%
  #2%
}
\makeatother

\DeclareCaptionStyle{ruled}{labelfont=normalfont,labelsep=colon,strut=off} 
\floatstyle{ruled}
\newfloat{listing}{tb}{lst}{}
\floatname{listing}{Listing}
\title{Two Heads Are Better than One: Distilling Large Language Model Features into Small Models with Feature Decomposition and Mixture}
\author {
    Tianhao Fu\equalcontrib\thanks{Work done when Tianhao Fu at Peking university.},
    Xinxin Xu\equalcontrib,
    Weichen Xu,
    Jue Chen,
    Ruilong Ren,\\
    Bowen Deng,
    Xinyu Zhao,
    Jian Cao\thanks{Corresponding author.},
    Xixin Cao
}
\affiliations {
    Peking University\\
    tianhaofu1@gmail.com, xuxinxin@stu.pku.edu.cn, caojian@ss.pku.edu.cn
}

\usepackage{bibentry}

\begin{document}

\maketitle

\begin{abstract}
Market making (MM) through Reinforcement Learning (RL) has attracted significant attention in financial trading. With the development of Large Language Models (LLMs), more and more attempts are being made to apply LLMs to financial areas. A simple, direct application of LLM as an agent shows significant performance. Such methods are hindered by their slow inference speed, while most of the current research has not studied LLM distillation for this specific task. To address this, we first propose the normalized fluorescent probe to study the mechanism of the LLM's feature. Based on the observation found by our investigation, we propose Cooperative Market Making (CMM), a novel framework that decouples LLM features across three orthogonal dimensions: layer, task, and data. Various student models collaboratively learn simple LLM features along with different dimensions, with each model responsible for a distinct feature to achieve knowledge distillation.
Furthermore, CMM introduces an H\'{a}jek-MoE to integrate the output of the student models by investigating the contribution of different models in a kernel function-generated common feature space. Extensive experimental results on four real-world market datasets demonstrate the superiority of CMM over the current distillation method and RL-based market-making strategies. 
\end{abstract}

%

\begin{figure}[t]
\centering
\includegraphics[width=\linewidth]{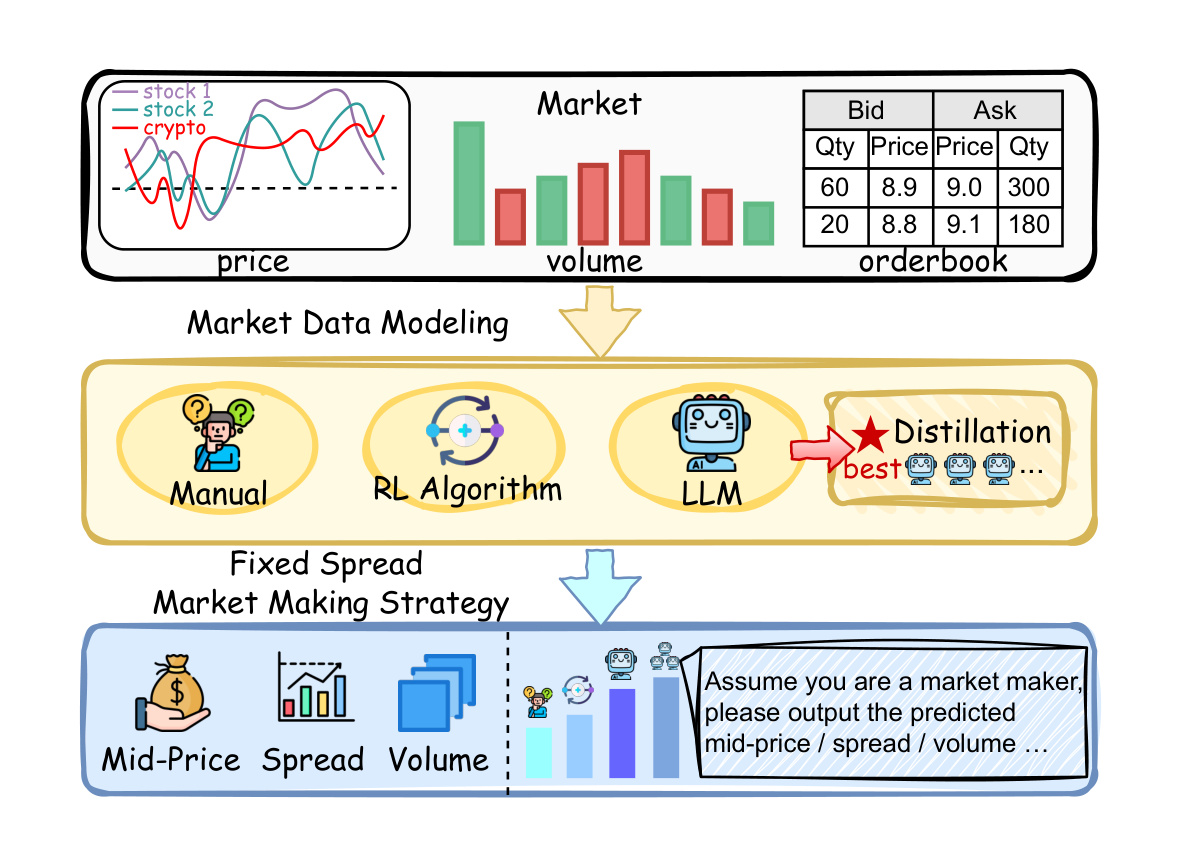}
\caption{Overview of the market-making workflow. A standard market-making algorithm analyzes historical market data and outputs future ordering strategies. We do a simple experiment that utilizes an LLM prompted with input to directly predict the future mid-price, spread, and volume, which is used to construct future orders via classic price and volume arithmetic sequences. We find that the LLM-based approach surpasses the performance of traditional RL algorithms. Furthermore, with our proposed distilled method, the small model demonstrates further significant improvements that could be used in a real-time scenario.}
\label{introduction_new}
\end{figure}
\section{Introduction}
Market making is a core task in financial area, which could provide liquidity to each financial asset \cite{gueant2013dealing}. Today, algorithmic systems handle more than 60\% of the trading volume in active markets \cite{othman2012automated}. Recent breakthroughs in LLM have demonstrated exceptional potential in financial data analysis \cite{brown2020language, radford2019language, li2025m2iv}. We conducted an exploratory experiment that shows Gemini-2.5-Pro \cite{team2023gemini} and Llama-3.1 \cite{grattafiori2024llama} perform well in all indicators, surpassing traditional RL methods, as depicted in Figure~\ref {introduction_new}. However, LLM inference speeds cannot meet the demands of real-time trading, where subsecond latency is essential \cite{vaswani2017attention, kaplan2020scaling}. Knowledge distillation (KD) \cite{hinton2015distilling, gou2021knowledge, xu2024survey} is a technique used to mitigate latency issues. However, existing approaches focus on distilling knowledge from large LLMs to small LLMs, these small LLMs still exhibit slower inference speeds compared to traditional lightweight models \cite{jiao2019tinybert}. There are no cross-architecture distillation methods. 


We then conduct an exploratory experiment that directly distills the LLM feature to a small model using traditional distillation methods, resulting in poor performance. This failure results in the single small model lacking the representational capacity to capture an LLM's deep and high-dimensional complex features. Therefore, we try to decompose complex LLM features into simpler components and assign them to an ensemble of small learning models to improve the effectiveness of small models in learning LLM features. We believe that layers, tasks, and input data types, such as these three variables, have the potential to decouple complex features. To further validate this, we propose a Normalized Fluorescent Probe to analyze the complex feature representations of LLMs \cite{yu2024interpreting}. Our analysis shows that with stronger decoupling conditions of these dimensions,
the LLM feature exhibits more apparent separation between clusters. Furthermore, a specialization across the model depth is observed: the shallow layers prioritize the prediction of mid-price, the middle layers focus on the spread, and the deep layers are geared towards the total volume\cite{bouchaud2009markets, kyle1985continuous}. Building on these insights, we introduce \textbf{ Cooperative Market Making (CMM)}, a two-stage framework which contains: (1) \textbf{ Orthogonal Feature Decomposition Distillation (OFDD)}: We decouple LLM features into three dimensions: layer feature hierarchy, task objectives, and data type/market regime to decompose complex feature spaces into specialized clusters \cite{lecun2015deep, tang2024divide}. Each cluster is distilled into dedicated lightweight models \cite{ba2014deep}. (2) \textbf{H\'{a}jek Projection-based Mixture-of-Experts (H\'{a}jek-MoE)}: We design a projection mechanism to quantify the contribution of each lightweight expert model. \cite{hajek1968asymptotic}


In summary, the contributions of this paper are as follows:
\begin{itemize}
    \item 
    We propose a Normalized Fluorescent Probe to perform a mechanistic analysis of LLM features and reveal two critical LLM features: (1) Features of different layers govern different outputs of the tasks. (2) Features within the same layer exhibit significant discrepancies when processing heterogeneous input data. 
    \item 
    We propose Orthogonal Feature Decomposition Distillation (OFDD) to decompose LLM features along with three complementary variables: (1) layer hierarchy, (2) task specialization, and (3) data market regime. This decomposition simplifies the LLM features and enables small models to learn from the LLM more effectively.
    \item 
    We propose the Hájek projection-based Mixture-of-Experts (Hájek-MoE) to integrate the outputs of different small models. Hájek-MoE quantifies the contribution of each model, considering the projection length of each model vector in the same feature space generated by a kernel function.
    \item 
    We demonstrate the superiority of our approach through extensive experiments conducted in challenging market environments. In contrast to LLM-based methods, our approach achieves higher accuracy and reduced computational cost. Compared to RL-based methods, it exhibits a much greater sample efficiency.
\end{itemize}

\section{Related Works}
\subsection{Market Making}
Market making involves the continuous submission of buy and sell orders in the limit order book \cite{gould2013limit} to maximize returns while managing risk \cite{vicente2023automated, amihud1980dealership}. Traditional MM frameworks, including those introduced by \cite{avellaneda2008high, gueant2017optimal}, are based on mathematical models that often assume static market conditions. In recent years, RL has gained traction as a flexible and adaptive approach for designing MM strategies that can respond to real-time market fluctuations \cite{zhong2021data, fang2021universal, lei2025large, sun2023mastering, 11175205, DBLP:conf/mm/Wu0YHFJYL24, zhao2025agentcdmenhancingmultiagentcollaborative}, but a significant portion of this research has focused on the refinement of strategies for a single price level \cite{sadighian2019deep}. Many RL-based approaches use the mid-price as a dynamic pricing reference, while this reliance on a volatile signal can lead to excessive order cancellations \cite{kumar2020deep}.

\subsection{LLM Distillation}
Recent LLMs, such as PaLM 540b \cite{chowdhery2023palm, li2025cama}, pose significant challenges to both inference and fine-tuning due to their substantial computational demands. These dependencies highlight the importance of knowledge distillation. Furthermore, the Chain-of-Thought \cite{wei2022chain} framework has enabled the generation of rich reasoning outputs from teacher models \cite{ho2022large}, allowing student models to learn not only the answers but also the reasoning processes. This approach enhances the student model through multitask learning \cite{chu2023multi}. In addition, efforts have been made to generate various rationales to improve the consistency of predictions \cite{chen2024towards}. Although these systems are designed to leverage knowledge diversity, they remain underexplored.

\subsection{Mixture of Expert}
Mixture of Experts (MoE) was introduced by \cite{jordan1994hierarchical}, with distinct experts handling different input sections. It was extended by deep MoE and conditional computation \cite{bertsekas2005convergence}. \cite{shazeer2017outrageously} integrated MoE with LSTMs \cite{hochreiter1997long}, and Switch Transformers \cite{fedus2022switch} combined MoE with transformers, enabling MoE in architectures like MoE-LLaVA \cite{lin2024moe}. The core principle of MoE is to expand the capacity of the model efficiently \cite{dou2019dynamic}. Unlike previous MoE techniques \cite{cai2024survey, vats2024evolution} where specialization emerges during training, MoAI \cite{lee2024moai} explicitly defines experts for specific input segments. Recent advances enable training trillion-parameter models in natural language processing \cite{li2021semantics} and computer vision \cite{ochs2015iteratively}.

\begin{figure*}[t]
\centering
\includegraphics[width=1\linewidth]{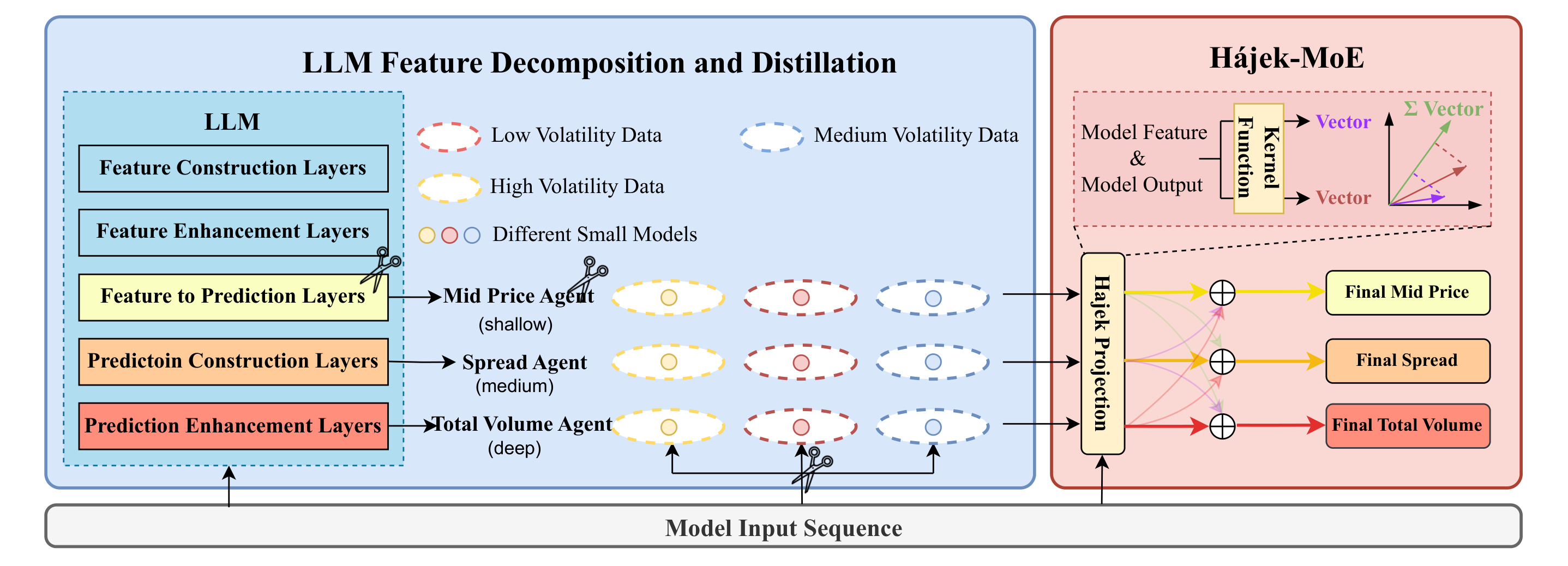}
\caption{Overview of the CMM Framework. Left: LLM Feature Decomposition and Distillation. The complex feature space of an LLM is decomposed across three dimensions: layer, task, and data. Such three variables result in various types of features, where each feature type is learned by a specialized small model, thereby effectively representing the comprehensive LLM feature space through a collection of smaller models. Right: Inference with Hájek-MoE. Hájek-MoE employs a kernel function to project the output and feature of each small model into a shared feature space to obtain each model's confidence score. The final prediction is computed by aggregating each model's output with the scores.}
\label{overview}
\end{figure*}

\section{Method}
In this section, we describe the proposed normalized fluorescent probe, the orthogonal feature decomposition distillation, and Hájek-MoE. Figure~\ref {overview} is an overview of our whole framework. Firstly, we decouple LLM features across three orthogonal dimensions: layer, task, and data. Various student models collaboratively learn complex LLM features along these dimensions, and each model is responsible for a specific type of feature. Next, through H'ajek-MoE, we integrate the output of the student models by investigating the contribution of the projection of different models along different input data vectors.

\subsection{Normalized Fluorescent Probe}
{
\parfillskip=0pt 
To study the LLM feature, we propose a normalized fluorescent probe that could accurately identify and quantify the influence of any position feature on the model's outputs. The normalized fluorescent probe follows a traditional LLM mechanistic interpretation approach that leverages noise perturbation techniques to assess the sensitivity and robustness of each feature and analyze the resultant changes in outputs. However, the traditional approach is easily affected by noise distributions, amplitudes, and output variation metrics. For example, when we study which feature has a greater impact on the specific task output, the results observed are different when adding Gaussian-distributed noise and adding uniformly distributed noise. Therefore, we proposed a noise-normalized probe that can ensure the robustness and comprehensiveness of our probe by doing multiple experiments with different noise distributions and averaging the results.

We compare perturbation effects within each module to identify its most influenced outputs. We ensure the robustness of the results by integrating different types of noise. The implementation steps of the normalized fluorescent probe are as Algorithm ~\ref{alg:nrom_probe}. As shown in lines 4 to 13, we first generate Gaussian and uniformly distributed noise and randomly sample each noise within a given amplitude range $[a_{min}, a_{max}]$ to construct noise samples with different amplitudes and distributions. Subsequently, we inject these noise samples into the parameters $\theta$ of the module one by one, as can be seen in line 14. By measuring and recording
\par 
}
\begin{algorithm}[H]
\caption{Normalized Fluorescent Probe}
\label{alg:nrom_probe}
\begin{algorithmic}[1]
\item[\textbf{Input:}] Model parameters $\theta$, feature set $F$, output variables $O$, 
noise types $\{\mathcal{N}, \mathcal{U}\}$, amplitude ranges $A$. 
\item[\textbf{Output:}] Causal Attribution Map $C$ (each element of $C$ is the index of the corresponding most influential module).

\FOR{each module $m \in M$}
    \STATE $\theta_{original} \gets \theta$ \COMMENT{Preserve original parameters}
    
    \STATE \textbf{1. Normalize over Noise Distributions}
    \FOR{each distribution $d \in \{\mathcal{N}, \mathcal{U}\}$}
        \IF{$d = \text{Gaussian}$}
            \STATE $\epsilon \sim \frac{1}{\sigma\sqrt{2\pi}} e^{-\frac{(x-\mu)^2}{2\sigma^2}}$ \COMMENT{$\mu=0, \sigma=1$}
        \ELSIF{$d = \text{Uniform}$}
            \STATE $\epsilon \sim  \frac{1}{b-a}  -1 \leq x \leq 1$ \\
        \ENDIF
    \ENDFOR
    \STATE \textbf{2. Normalize over Output Amplitude}
    \FOR{each amplitude range $[a_{min}, a_{max}] \in A$}
        \STATE $A_{scale} \gets \text{random}(a_{min}, a_{max})$
        \ColorState{myhighlightcolor}
        {$\theta_{perturbed} \gets \theta + A_{scale} \cdot \epsilon$}
        \FOR{each output $o \in O$}
            \ColorState{myhighlightcolor}
            {$\Delta_o = |f(\theta_{perturbed})_o - f(\theta_{original})_o|$}
            \STATE $\Delta_{norm} \gets \frac{\Delta_o - a_{min}}{a_{max} - a_{min}}$
    \ENDFOR
        \STATE $S_{m,o} \gets \frac{1}{|A|\cdot N}\sum_{i=1}^{N} \Delta_{norm}^{(i)}$
    \ENDFOR
\ENDFOR
\STATE $C_o \gets \operatorname*{arg\,max}_{m \in M} S_{m,o}$ for each output $o \in O$ \COMMENT{Find the module with max influence for each output}
\STATE \textbf{return} Causal Attribution Map $C$
\end{algorithmic}
\end{algorithm}
\noindent
the output change values $\Delta_o$ after the changes of the corresponding parameters, we can evaluate the impact of the module on the output. As illustrated in lines 16 to 19, in order to eliminate the interference of noise amplitude differences and examine the effects of different noise forms, after the changes of the corresponding parameters, we can evaluate the impact of the module on the output. As illustrated in lines 16 to 19, in order to eliminate the interference of noise amplitude differences and examine the effects of different noise forms, we normalize the noise amplitude of each output $\Delta_o$ to obtain $\Delta_{norm}$ and calculate its average effect $\text{Avg}_{o}$ under different noise forms. Then we get an M×O matrix S, where the row index represents different modules, the column index represents a specific output, and each element in the matrix reflects the contribution or influence of the corresponding module on the output. The Causal Attribution Map $C$ is obtained by taking the maximum value of each column in the S matrix, which reveals the module with the greatest impact on each output.

\begin{figure}[t]
    \centering
    \includegraphics[width=\linewidth]{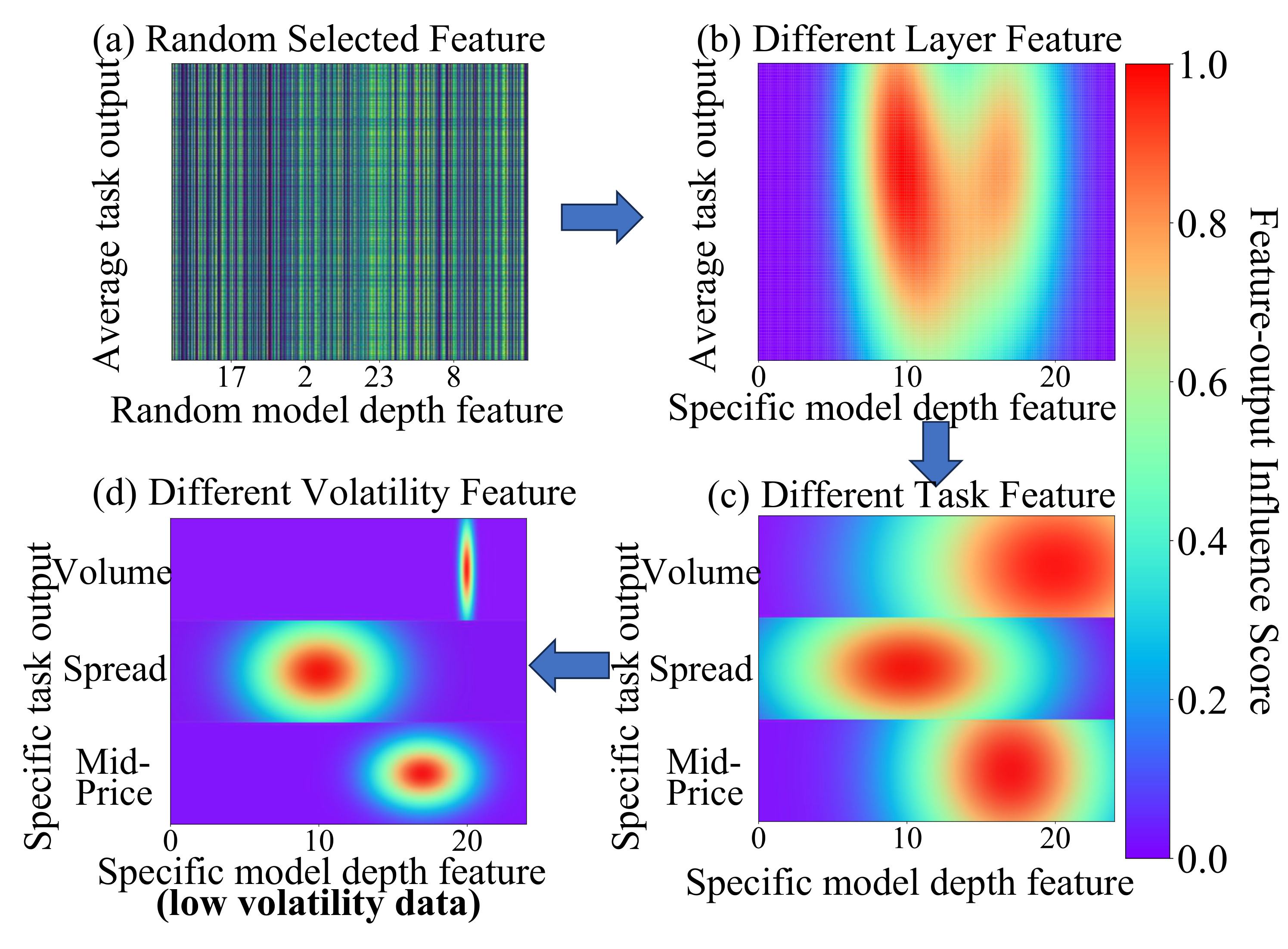}
    \caption{Progressive feature decomposition visualization by our probe results. With stronger decoupling conditions, the LLM features exhibit clearer separation between clusters. Furthermore, a specialization across model depth is observed: shallow layers prioritize mid-price prediction, middle layers focus on the spread, and deep layers are geared towards total volume.}
    \label{intro}
\end{figure}

After applying noise perturbations and computing the influence scores for each pair (module, output), we then quantify the impact of each module on each output.
Based on the normalized fluorescent probe, we find that in market making, shallow LLM features are most useful for predicting mid-price, middle features for spread, and deep features for total volume.
We also find that even in the same layer, representations exhibit significant discrepancies when processing heterogeneous input data. For example, feature clustering is more pronounced when visualizing data with low volatility. Figure ~\ref{intro} shows the visualization results.

\subsection{Orthogonal Feature Decomposition Distillation}
As mentioned above, our normalized fluorescent probe analysis reveals that LLM features can be classified along three dimensions. Therefore, we reduce the learning difficulty for small models by decomposing the complex features along these dimensions, and each class of features is distilled with a specific small model.

\subsubsection{Base Distillation.}
\begin{figure}[t]
    \centering
    \includegraphics[width=\linewidth]{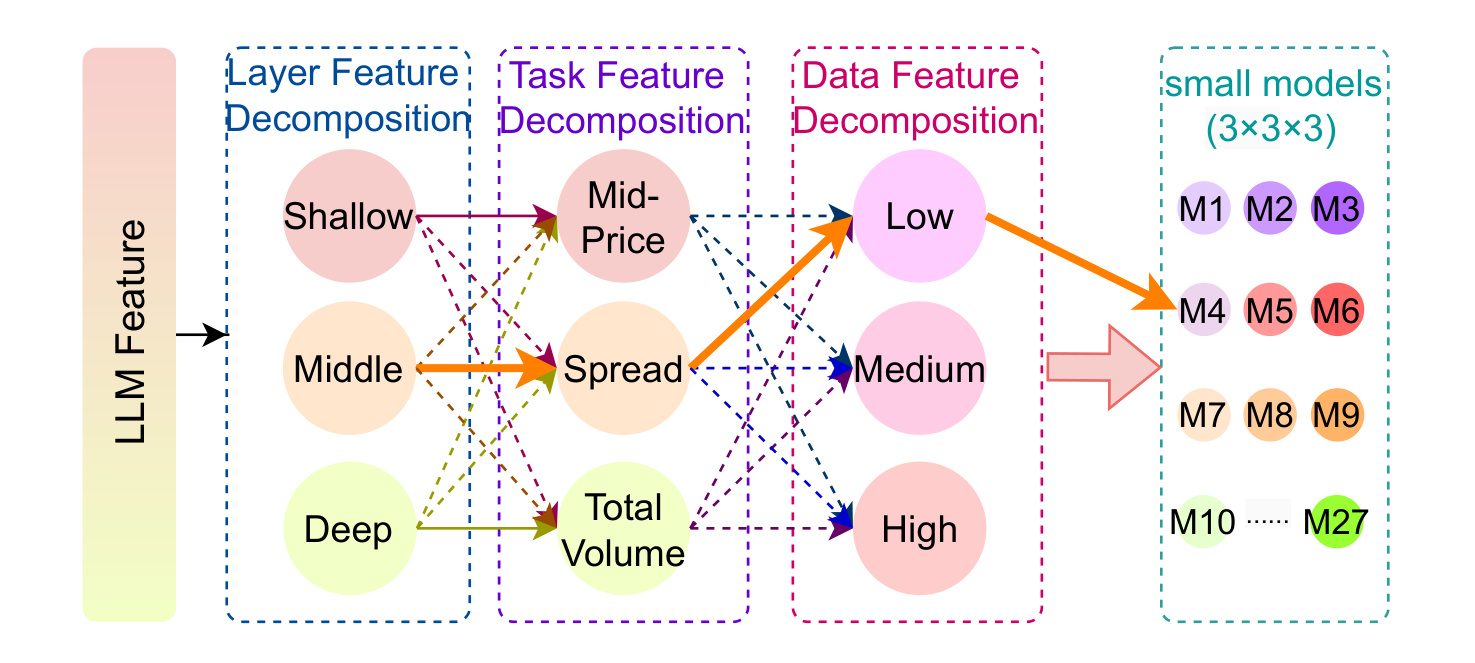}
    \caption{Tree diagram of Orthogonal Feature Decomposition Distillation. By varying three decoupling variables, complex LLM features are decomposed into simpler components and distilled into specialized small models.}
    \label{tree_method}
\end{figure}

We perform base distillation by distilling the features during the training process. However, our initial experiments did not show good results. This is because the small model's architecture is too simple to capture the complexity of the LLM's advanced features. Since we have shown above that the features of different modules of LLM mainly affect different outputs, we use various small models to learn these different features. In view of this, we propose a strategy to decouple the complex features of LLM from three dimensions, and thereby improve the feature representation capabilities of small models. Specifically, we decompose LLM features into three independent types, each of which is learned by a dedicated small model. Notably, because of the observation that a high degree of correlation exists between layer-wise and task-wise decompositions, for the layer and task variables, the corresponding pair distilled small model has a large weight when doing MoE. Figure~\ref{tree_method} shows the specific decomposition process. In this way, the features learned by multiple small models can effectively approximate the overall features of the LLM, significantly reducing the difficulty of training small models.

\subsubsection{Layer Feature Decomposition.}
Previous work has shown that the different layers of LLM are characterized at different levels, from shallow to deep \cite{yu2024interpreting}. Our normalized fluorescent probe also identified the dividing lines between different layers. Based on this intuition, we make different models learn different parts of the feature during Distillation, which can effectively reduce the difficulty of feature learning. In this way, we initially decompose the features based on the layer. The last three levels that we have discovered with the probe are the output of the separate controls, mid-price, spread, and total volume. 


\subsubsection{Task Feature Decomposition.}
In addition to decomposing the structure of the LLM, we can further enhance the distillation performance of the small model by decomposing our task. Our specific task involves determining the mid-price, the spread, and the total volume of market making orderbook.

Referring to Figure~\ref{intro}(c), we found that different LLM layers specialize in predicting different outputs. As mentioned in Layer Feature Decomposition, we identified that the LLM Feature to Prediction Layers (Shallow Layer), LLM Prediction Construction Layers (Middle Layer), and LLM Prediction Enhancement Layer (Deep Layer) exhibit the strongest correlations with the outputs mid-price, spread, and total volume, respectively. Leveraging these insights, the lightweight model, which aims to learn shallow features, will additionally learn the LLM mid-price output logits through both feature distillation and logit distillation. In this way, we further decompose the LLM features based on the task.

\subsubsection{Data(Market Type) Feature Decomposition.}
From the normalized fluorescent probe, we recognized that the features of the same layer exhibit significant discrepancies when processing input data with different market volatility, as demonstrated in Figure~\ref{intro}(d). Therefore, based on the input data, we calculate the historical 5-day volatility of each data sample and then categorize the input data into three types by dividing the volatility into three bands. Thus, we can further decompose the features according to the kind of data. Training a market maker model on low, medium, or high volatility data results in conservative, neutral, and aggressive strategies, respectively. In this way, we further decompose the LLM features based on the type of market.

To further improve the performance of small models, we also experimented with different architectures that learn the same feature type to identify the optimal ones.


\begin{table*}[t] 
    \centering
    \setlength{\tabcolsep}{4pt} 
    \renewcommand{\arraystretch}{1.2} 
    \resizebox{\textwidth}{!}{ 
    \begin{tabular}{lcccccccccccc}
        \hline
        \rowcolor[HTML]{F7E7CE}
         & \multicolumn{3}{c}{\textbf{RB}} & \multicolumn{3}{c}{\textbf{FU}} & \multicolumn{3}{c}{\textbf{CU}} & \multicolumn{3}{c}{\textbf{AG}} \\
        \cline{2-13}
        \rowcolor[HTML]{F7E7CE}
         & \textbf{EPnL [$10^3$] $\uparrow$} & \textbf{MAP [unit] $\downarrow$} & \textbf{PnLMAP $\uparrow$} 
         & \textbf{EPnL [$10^3$] $\uparrow$} & \textbf{MAP [unit] $\downarrow$} & \textbf{PnLMAP $\uparrow$} 
         & \textbf{EPnL [$10^3$] $\uparrow$} & \textbf{MAP [unit] $\downarrow$} & \textbf{PnLMAP $\uparrow$} 
         & \textbf{EPnL [$10^3$] $\uparrow$} & \textbf{MAP [unit] $\downarrow$} & \textbf{PnLMAP $\uparrow$} \\
        \hline
        FOIC & 3.23 $\pm$ 4.35 & 255 $\pm$ 111 & 14 $\pm$ 22 & -7.79 $\pm$ 9.25 & 238 $\pm$ 135 & -43 $\pm$ 56 & -33.05 $\pm$ 27.63 & 206 $\pm$ 141 & -161 $\pm$ 224 & -48.39 $\pm$ 28.83 & 189 $\pm$ 154 & -250 $\pm$ 335 \\
        LIIC & 2.26 $\pm$ 3.32 & 123 $\pm$ 32 & 20 $\pm$ 29 & -6.89 $\pm$ 6.66 & 115 $\pm$ 30 & -66 $\pm$ 69 & -24.19 $\pm$ 14.83 & 150 $\pm$ 20 & -164 $\pm$ 513 & -38.9 $\pm$ 26.2 & 142 $\pm$ 45 & -302 $\pm$ 243 \\
        LTIIIC & 9.16 $\pm$ 4.87 & 65 $\pm$ 6 & 139 $\pm$ 68 & 8.26 $\pm$ 2.64 & 52 $\pm$ 3 & 160 $\pm$ 50 & -16.74 $\pm$ 15.81 & 112 $\pm$ 109 & -190 $\pm$ 203 & -32.57 $\pm$ 22.8 & 128 $\pm$ 22 & -264 $\pm$ 166 \\
        $RL_{SD}$ & 4.36 $\pm$ 1.64 & 38 $\pm$ 4 & 114 $\pm$ 38 & 7.31 $\pm$ 5.38 & 76 $\pm$ 29 & 90 $\pm$ 46 & -19.7 $\pm$ 17.2 & 214 $\pm$ 109 & -92 $\pm$ 298 & -25.43 $\pm$ 23.83 & 107 $\pm$ 37 & -237 $\pm$ 235 \\
        $DRL_{OS}$ & 8.22 $\pm$ 3.70 & 51 $\pm$ 4 & 156 $\pm$ 61 & 11.03 $\pm$ 13.87 & 37 $\pm$ 3  & 30 $\pm$ 36 & -18.9 $\pm$ 18.02 & 647 $\pm$ 2367 & -99 $\pm$ 147 & -28.39 $\pm$ 27.92 & 169 $\pm$ 154 & -167 $\pm$ 135 \\
        IMM & 16.46 $\pm$ 9.1 & 96 $\pm$ 13 & 165 $\pm$ 74 & 28.1 $\pm$ 10.27 & 102 $\pm$ 14 & 274 $\pm$ 89 & -4.86 $\pm$ 10.17 & 111 $\pm$ 28 & -43 $\pm$ 87 & -14.5 $\pm$ 20.2 & 102 $\pm$ 14 & -274 $\pm$ 89 \\
        \rowcolor{gray!10} 
        \textbf{CMM}    & \textbf{22.69 \boldmath$\pm$ 1.96} & \textbf{34 \boldmath$\pm$ 3} & \textbf{179 \boldmath$\pm$ 14}
                         & \textbf{31.39 \boldmath$\pm$ 4.18} & \textbf{32 \boldmath$\pm$2} & \textbf{298 \boldmath$\pm$ 19}
                         & \textbf{-1.63 \boldmath$\pm$ 0.41} & \textbf{30 \boldmath$\pm$ 8} & \textbf{-16 \boldmath$\pm$ 17}
                         & \textbf{-8.95 \boldmath$\pm$ 11.22} & \textbf{64 \boldmath$\pm$ 2} & \textbf{-143 \boldmath$\pm$ 102} \\
        
        \hline
    \end{tabular}
    }
    \caption{Overall Results.}
    \label{tab:comparison_results}
\end{table*}

\subsection{Hájek Projection-based Mixture-of-Experts}
After the OFDD, each small model could and only could capture a specific type of feature of the original LLM, resulting in poor performance when processing other types of data features. Therefore, we urgently need to find an effective way to combine these highly specialized small models. When choosing a model fusion strategy, we did not adopt the conventional MoE framework. Because our situation is unique: The small models that we built have unique architectures and inherent correlations, while they inherit specific knowledge at different levels from the same LLM. 


Drawing inspiration from the Hájek projection \cite{hajek1968asymptotic,wager2018estimation}, which utilizes a kernel function to map the different spaces into the same space, to manage and leverage various small models effectively, we employ a framework based on the mixture of experts (Hájek-MoE) based on the Hájek projection. Our approach similarly maps the feature spaces of different models into a unified representation space through a kernel function that accepts each model's features and prediction and outputs each model vector's coordinates in the mapped kernel space.

In practice, the Hájek Projection is realized through a concrete strategy of dimensionality reduction and geometric projection to compute the confidence score of each expert. The key steps are as follows:

First, we define a kernel function $\phi$, which maps high-dimensional expert features and predictions into a 2D vector space. In practice, $\phi$ is implemented as a shallow neural network.

Next, we compute the consensus feature by computing the average feature and prediction across all $N$ experts, denoted as $\bar{E}(X)$:
\begin{align}
\bar{E}(X) = \frac{1}{N} \sum_{j=1}^{N} E_j(X)
\end{align}
This average output is then mapped to a 2D vector using the function $\phi$, which we call the consensus vector $V_{\text{avg}}$. This vector serves as the reference axis for our projection.
\begin{align}
V_{\text{avg}} = \phi(\bar{E}(X))
\end{align}

Similarly, for each expert $E_i$, its output $E_i(X)$ is mapped to a 2D feature vector $V_i$ using the same function $\phi$.

The Hájek confidence score $C_i$ for each expert $E_i$ is defined as the scalar projection of its feature vector $V_i$ onto the consensus vector $V_{\text{avg}}$. This value quantifies the degree of alignment between the expert's contribution and the collective consensus. The confidence score is computed as:
\begin{align}
C_i = \frac{V_i \cdot V_{\text{avg}}}{\|V_{\text{avg}}\|}
\end{align}
where $V_i \cdot V_{\text{avg}}$ is the dot product of the two vectors, and $\|V_{\text{avg}}\|$ is the Euclidean norm of the consensus vector. This ensures that experts whose outputs are more aligned with the group consensus receive a higher Hájek confidence score.

\section{Experiments}
 
\subsection{Experimental Setup}

\subsubsection{Dataset and Setting.}
The data we use are historical orders and trades on the Shanghai Futures Exchange from July 2021 to July 2022, covering 186 trading days follows IMM \cite{niu2023imm}. We select four contracts as 4 datasets: \textit{FU}, \textit{RB}, \textit{CU}, and \textit{AG} and reconstruct their historical 5-depth limit order books with a 500-millisecond real-time financial period. Our model takes the order book of the previous timestep as input. It directly predicts the mid-price, spread, and volume for the subsequent timestep, which are then used to construct the new order book via a traditional trapezoidal algorithm. We follow the IMM \cite{niu2023imm} to do the dataset split to train and test. We use LLaMA3.1 as our LLM. We use MLP with 2 layers as hajak kernel function. All experiments are conducted on 64 NVIDIA H100 GPUs. \\
\subsubsection{Compared Methods and Metrics.}
We evaluate the effect of our method on multiple rule-based KD approaches, including FOIC \cite{gavsperov2021market}, LIIC \cite{gavsperov2021market}, LIIIC \cite{niu2023imm}. We also compare with various RL-based MM methods including $RL_{DS}$ \cite{beysolow2019market}, $DRL_{OS}$ \cite{chung2022market}, IMM \cite{niu2023imm} and CMM. The following performance and risk metrics are employed: 1.Episodic PnL is a natural choice to evaluate the profitability of a MM agent \cite{sutton2018reinforcement}. 2.Mean Absolute Position (MAP) accounts for the inventory risk \cite{garleanu2013dynamic}. 3.Return Per Trade (RPT) evaluates the agent’s capability of capturing the spread \cite{easley2011microstructure}. It is normalized across different markets by the average market spread. 4. PnLMAP defined as PnL divided by the mean absolute position (MAP) in this period \cite{hull2016options}. It means the PnL in per unit of inventory and can measure the ability of the agent to profit against inventory risk.

\subsection{Overall Results}
The experimental results in Table~\ref{tab:comparison_results} demonstrate the superior performance of CMM in all 4 futures contracts. For instance, on the RB dataset, CMM not only achieves significantly higher profitability than the strongest baseline IMM, but also substantially reduces inventory risk. This was accomplished through layer-task aligned distillation guided by the normalized fluorescent probe. The FU contract results further validate the framework's robustness, where our method attains the highest PnLMAP of 298, a significant improvement over IMM. This superior performance is achieved by effectively integrating volatility-conditioned specialization, thereby demonstrating stable profitability across varying market regimes. Notably, in challenging low-liquidity markets such as CU and AG contracts, CMM significantly outperforms the benchmarks in terms of the terminal wealth, risk-adjusted return, and spread-capturing ability. This performance stems from Hájek-MoE's adaptive fusion mechanism, which dynamically obtains the confidence scores of experts based on the real-time market. 

\subsection{Model Component Ablation Study}
Table~\ref{tab:Ablation Study on FU Dataset} demonstrates the progressive impact of our decomposition components. SR refers to the Sharpe Ratio. The layer-wise decomposition baseline, FD(layer), delivers the poorest performance across all metrics. This result indicates that only isolating hierarchical LLM features is insufficient. An improvement in profitability is achieved by introducing task decomposition with specialized prediction heads. This enhancement arrives from the model's newfound ability to develop distinct, focused strategies for different predictive sub-tasks, moving beyond a monolithic prediction approach and thus capturing market opportunities more effectively. Subsequently, volatility-type decomposition improves risk control through market-regime adaptation, and the model successfully mitigates potential losses, as reflected by a marked reduction in adverse outcomes. 
The complete framework, FD(Hájek-MoE), achieves optimal balance between profitability and risk-adjusted returns, which demonstrates the superiority of its input-adaptive fusion mechanism. Unlike static aggregation, our approach dynamically allocates confidence scores to the experts in real-time, achieving a robust fusion of all decomposed components.

\begin{table}[h!]
    \centering
    \setlength{\tabcolsep}{4pt}
    \renewcommand{\arraystretch}{1.2}
    \resizebox{0.48\textwidth}{!}{
    \begin{tabular}{lcccc}
        \hline
        \rowcolor[HTML]{F7E7CE}
        \textbf{Models} & \textbf{EPnL [$10^{3}$] $\uparrow$} & \textbf{MAP [unit] $\downarrow$} & \textbf{PnLMAP $\uparrow$} & \textbf{SR $\uparrow$} \\
        \hline
        CMM$_{\text{FD(layer)}}$ & 12.54 $\pm$ 3.64 & 67 $\pm$ 5 & 162 $\pm$ 46 & 1.98 \\
        CMM$_{\text{FD(task)}}$ & 17.25 $\pm$ 9.72 & 58 $\pm$ 47 & 195 $\pm$ 56 & 2.33 \\
        CMM$_{\text{FD(type)}}$ & 22.41 $\pm$ 5.11 & 45 $\pm$ 5  & 241 $\pm$ 57 & 2.68 \\
        \rowcolor{gray!10} 
        CMM$_{\text{FD(hájek-moe)}}$ & \textbf{31.39 \boldmath$\pm$ 4.18} & \textbf{32 \boldmath$\pm$ 2} & \textbf{298 \boldmath$\pm$ 19} & \textbf{2.98} \\
        \hline
    \end{tabular}
    }
    \caption{Comparison results of ablation study on FU dataset.}
    \label{tab:Ablation Study on FU Dataset}
\end{table}

\subsection{Comparison of Different Distillation Methods and MoE Methods }
As shown in Table~\ref{tab:LLM componment}, our framework achieves a higher EPnL of 31.39, an improvement of 3.02\% compared to the original LLM, while also demonstrating 6.3× lower latency at just 0.3s. This validates the effectiveness of our orthogonal decomposition strategy in preserving critical features identified by the normalized fluorescent probe. Conventional KD methods such as ReviewKD and CAT-KD suffer severe performance degradation due to entangled feature learning, while standard MoE methods like X-MoE and MH-MoE exhibit poor risk control from static expert fusion. CMM's superiority stems from its three-dimensional decomposition.

\begin{table}[h!]
    \centering
    \setlength{\tabcolsep}{2pt}
    \renewcommand{\arraystretch}{1.2}
    \resizebox{0.48\textwidth}{!}{
    \begin{tabular}{lccccc} 
        \hline
        \rowcolor[HTML]{F7E7CE}
        \textbf{Models} & \textbf{EPnL [$10^3$] $\uparrow$} & \textbf{MAP [unit] $\downarrow$} & \textbf{PnLMAP $\uparrow$} & \textbf{SR $\uparrow$} & \textbf{Latency(s) $\downarrow$} \\
        \hline
        LLM-Base     & 30.47 $\pm$ 5.52   & 64 $\pm$ 13   & 283 $\pm$ 66  & 3.01 $\pm$ 0.3 & 1.9 \\
        ReviewKD     & 10.52 $\pm$ 3.21   & 2230 $\pm$ 112  & 48 $\pm$ 9    & 2.00 $\pm$ 0.0 & 0.22 \\
        Sim-KD       & 15.34 $\pm$ 4.15   & 1895 $\pm$ 85   & 85 $\pm$ 14   & 2.25 $\pm$ 0.1 & 0.22 \\
        CAT-KD       & 20.81 $\pm$ 5.03   & 1570 $\pm$ 143  & 132 $\pm$ 18  & 2.48 $\pm$ 0.2 & 0.22 \\
        X-MoE        & 19.63 $\pm$ 6.17   & 1180 $\pm$ 205  & 195 $\pm$ 23  & 1.67 $\pm$ 0.3 & 0.46 \\
        MH-MoE       & 20.92 $\pm$ 3.52   & 785 $\pm$ 315   & 155 $\pm$ 28  & 1.89 $\pm$ 0.1 & 0.57 \\
        \rowcolor{gray!10} 
        \textbf{CMM} & \textbf{31.39 \boldmath$\pm$ 4.18} & \textbf{32 \boldmath$\pm$ 2} & \textbf{298 \boldmath$\pm$ 19} & \textbf{2.98} & \textbf{0.3} \\
        \hline
    \end{tabular}
    }
    \caption{Comparison of Different Distillation Methods and MoE Methods on FU.}
    \label{tab:LLM componment}
\end{table}


\subsection{Feature Decomposition Analysis}
\begin{figure}[t]
\centering
\begin{subfigure}[t]{0.49\columnwidth} 
    \centering
    \includegraphics[width=\linewidth]{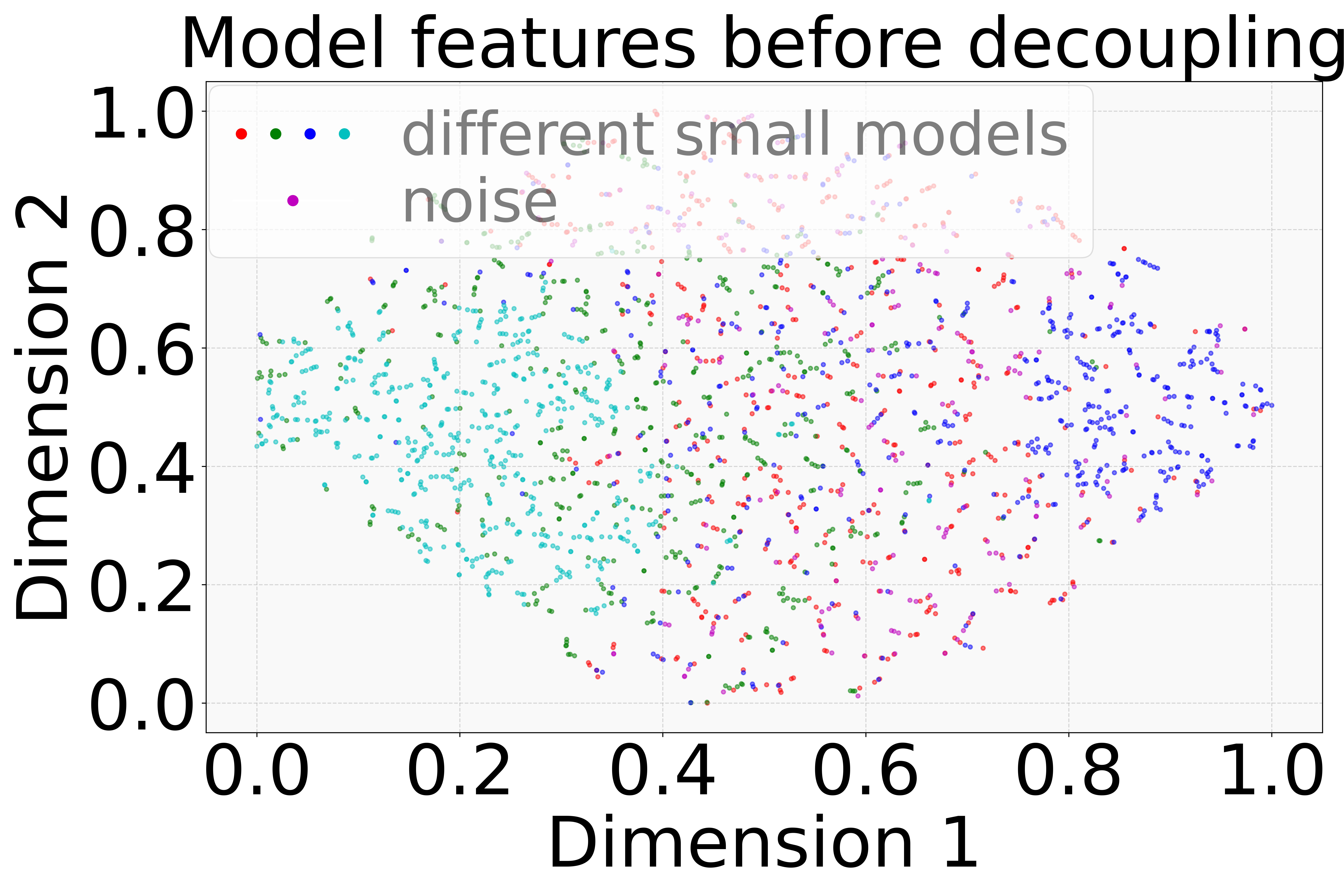}
    \label{subfig1}
\end{subfigure}
\begin{subfigure}[t]{0.49\columnwidth}
    \centering
    \includegraphics[width=\linewidth]{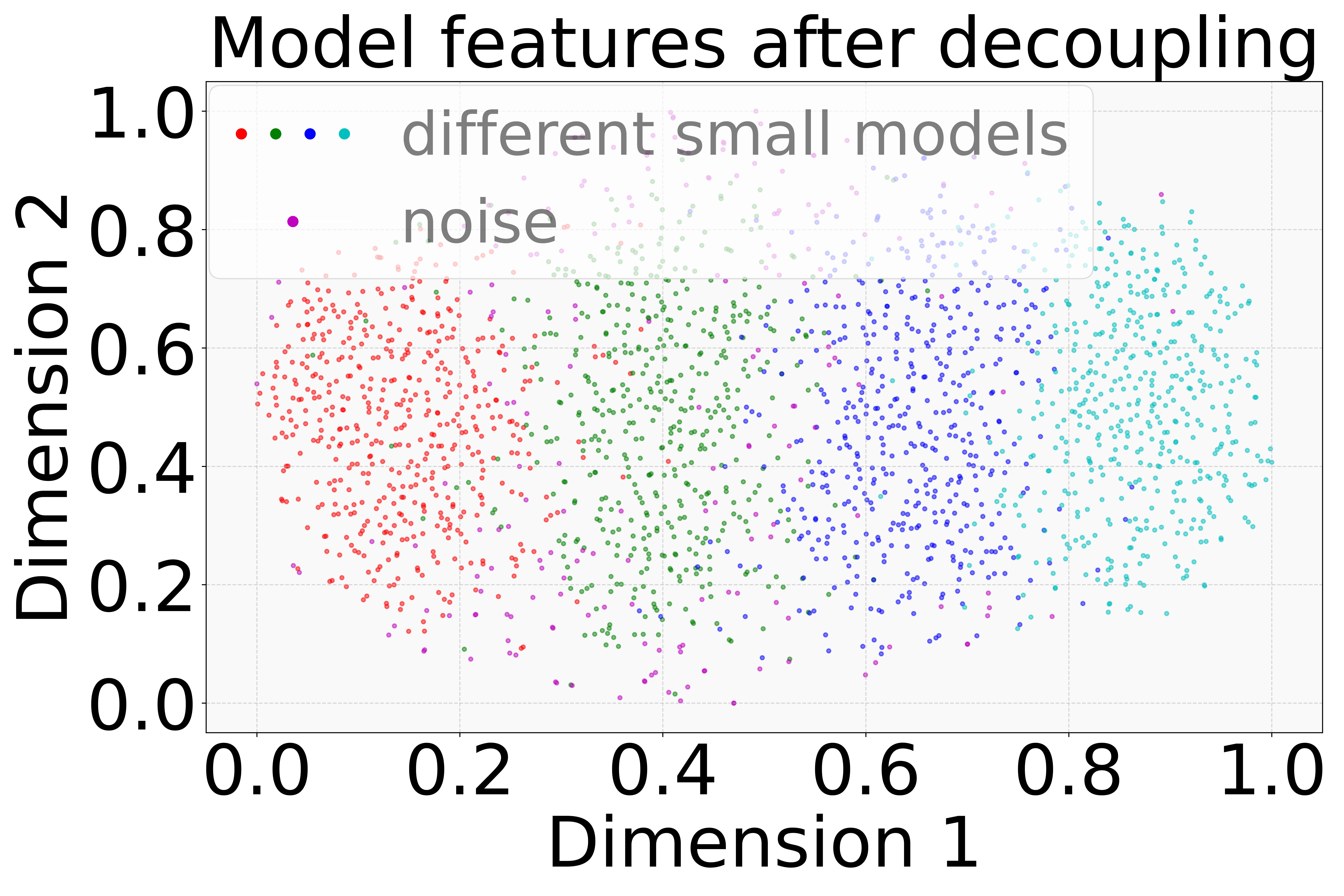}
    \label{subfig2}
\end{subfigure}

\caption{Feature Decomposition Analysis}
\label{FDA}
\end{figure}

Figure~\ref{FDA} presents a 2D visualization of the feature space, obtained via PCA, to demonstrate the effectiveness of orthogonal decomposition. Each point in the figure corresponds to the feature representation from a small model. The pre-decomposition features are chaotically overlapped. In contrast, the post-decomposition features reveal a highly organized structure. This structure is visually articulated through color, where models sharing the same volatility, task, and layer depth are assigned the same color and visibly cluster together. For example, all models for the spread task in a low-volatility regime using a shallow layer appear as a distinct green cluster, while models for the total volume task in a high-volatility regime with a deep layer form a blue cluster. 

\subsection{Robustness Test Under Extreme Market Conditions}

To evaluate the robustness of CMM under extreme market, we conducted experiments using historical data augmented with simulated flash crash and sudden reversal scenarios.

The results in Table \ref{tab:robustness} show that CMM achieves a strong balance of profitability and risk mitigation under duress, where baseline models proved vulnerable. This resulted in a healthier risk-adjusted return. These results demonstrate that CMM is more robust under extreme market, likely due to its dynamic expert confidence score mechanism, which prioritizes risk control experts during market turbulence.

\begin{table}[h!]
\centering
\small 
\setlength{\tabcolsep}{4pt} 
\resizebox{0.48\textwidth}{!}{
\begin{tabular}{lcccc}
\hline
\rowcolor[HTML]{F7E7CE}
\textbf{Models} & \textbf{EPnL [$10^3$] $\uparrow$} & \textbf{MAP (unit)} $\downarrow$ & \textbf{PnLMAP} $\uparrow$ & \textbf{SR} $\uparrow$ \\ \hline
LLM-Base       & 6.50 $\pm$ 2.45         & 150 $\pm$ 30             & 0.43 $\pm$ 0.08       & 0.60 $\pm$ 0.1       \\
IMM           & 7.80 $\pm$ 3.05         & 120 $\pm$ 25             & 0.65 $\pm$ 0.12       & 0.80 $\pm$ 0.1       \\
\rowcolor{gray!10} 
\textbf{CMM}  & \textbf{10.50 \boldmath$\pm$ 2.10}          & \textbf{45 \boldmath$\pm$ 8}              & \textbf{2.33 \boldmath$\pm$ 0.46}       & \textbf{1.20 \boldmath$\pm$ 0.2}       \\ \hline
\end{tabular}
}
\caption{Robustness Test Results under Extreme Market.}
\label{tab:robustness}
\end{table}

\subsection{Long-Term Market Adaptability Experiment}
To assess the long-term adaptability of CMM across different market regimes, we ran the model over a 1-week period covering various market conditions, including bull, bear, and sideways markets. The results in Table \ref{tab:long-term} show that CMM achieves superior performance compared to LLM-Base and IMM in terms of profitability, risk control, and computational efficiency across different market regimes. These results indicate that CMM is better at adapting to changing market, likely due to its orthogonal decomposition strategy, which allows CMM to specialize in different market types.

\begin{table}[h!]
\centering
\resizebox{0.48\textwidth}{!}{
\setlength{\tabcolsep}{4pt} 
\begin{tabular}{lcccc}
\hline
\rowcolor[HTML]{F7E7CE}
\textbf{Models} & \textbf{EPnL [$10^3$] $\uparrow$} & \textbf{MAP (unit)} $\downarrow$ & \textbf{PnLMAP} $\uparrow$ & \textbf{SR} $\uparrow$ \\ \hline
LLM-Base       & 12.00 $\pm$ 1.50  & 80 $\pm$ 15  & 0.56 $\pm$ 0.12 & 1.80 $\pm$ 0.20 \\
IMM            & 10.00 $\pm$ 1.00  & 60 $\pm$ 10  & 0.45 $\pm$ 0.10 & 1.50 $\pm$ 0.10 \\
\rowcolor{gray!10} 
\textbf{CMM}  & \textbf{14.00 \boldmath$\pm$ 1.20}       & \textbf{30 \boldmath$\pm$ 3}       & \textbf{0.80 \boldmath$\pm$ 0.05} & \textbf{2.20 \boldmath$\pm$ 0.15} \\ \hline
\end{tabular}
}
\caption{Long-Term Market Adaptability Results.}
\label{tab:long-term}
\end{table}

\subsection{Performance Test Under Low-Data Conditions}
To evaluate the performance of CMM when trained with limited data, we conducted experiments using 10\%, 20\%, and 50\% of the available data. The results in Table \ref{tab:low-data} show that CMM consistently outperforms the baselines across all data percentages. Even when trained on a minimal fraction of the data, CMM establishes a commanding lead by delivering markedly higher returns while simultaneously maintaining much stricter risk control. This may be because baseline models are prone to overfitting or failing to discern clear patterns from limited information. CMM's architecture allows it to learn more generalizable market behaviors. This data efficiency can likely be attributed to its orthogonal decomposition strategy, which simplifies the feature space and reduces the complexity of the learning task.

\begin{table}[h!]
\centering
\setlength{\tabcolsep}{4pt} 
\resizebox{0.48\textwidth}{!}{
\begin{tabular}{lcccc}
\hline
\rowcolor[HTML]{F7E7CE}
\textbf{Data \% } & \textbf{Models} & \textbf{EPnL [$10^3$] $\uparrow$} & \textbf{MAP (unit)} $\downarrow$ & \textbf{PnLMAP} $\uparrow$ \\ \hline
10\%                     & LLM-Base       & 2.50 ± 0.85         & 80 ± 15               & 0.31 ± 0.06       \\
& IMM            & 2.75 ± 0.95         & 52 ± 8                & 0.53 ± 0.10       \\
\rowcolor{gray!10} 
& \textbf{CMM}  & \textbf{4.50 \boldmath± 1.20} & \textbf{35 \boldmath± 5}       & \textbf{1.29 \boldmath± 0.26} \\ \hline
20\%                     & LLM-Base       & 3.20 ± 1.05         & 72 ± 12               & 0.44 ± 0.08       \\
& IMM            & 3.50 ± 1.10         & 48 ± 7                & 0.73 ± 0.14       \\
\rowcolor{gray!10} 
& \textbf{CMM}  & \textbf{5.20 \boldmath± 1.30} & \textbf{32 \boldmath± 4}       & \textbf{1.63 \boldmath± 0.32} \\ \hline
50\%                     & LLM-Base       & 4.00 ± 1.20         & 65 ± 10               & 0.62 ± 0.12       \\
& IMM           & 4.30 ± 1.05         & 40 ± 6                & 1.08 ± 0.21       \\
\rowcolor{gray!10} 
& \textbf{CMM}  & \textbf{6.00 \boldmath± 1.40} & \textbf{28 \boldmath± 3}       & \textbf{2.14 \boldmath± 0.43} \\ \hline
\end{tabular}
}
\caption{Performance under Low-Data Conditions.}
\label{tab:low-data}
\end{table}

\subsection{Energy Efficiency and Inference Speed Experiment}
To compare the energy efficiency and inference speed of CMM with baseline methods, we measured power consumption and inference latency during model deployment. The results in Table \ref{tab:efficiency} show that CMM significantly outperforms the baselines in terms of energy efficiency and inference speed. These results indicate that CMM is more suitable for real-time trading environments, where low latency and energy efficiency are critical.

\begin{table}[h!]
\centering
\setlength{\tabcolsep}{5pt} 
\resizebox{0.48\textwidth}{!}{
\begin{tabular}{lccc}
\hline
\rowcolor[HTML]{F7E7CE}
\textbf{Models} & \textbf{Power Usage (W)} ↓ & \textbf{Inference Latency (ms)} ↓ \\ \hline
LLM-Base       & 150 $\pm$ 10             & 120 $\pm$ 15       \\
IMM            & 90 $\pm$ 5               & 80 $\pm$ 10        \\
\rowcolor{gray!10} 
\textbf{CMM}  & \textbf{45 \boldmath$\pm$ 3}      & \textbf{20 \boldmath$\pm$ 5}  \\ \hline
\end{tabular}
}
\caption{Energy Efficiency and Inference Speed Results.}
\label{tab:efficiency}
\end{table}

\section{Conclusion}
This paper addresses market-making challenges by proposing CMM, a framework that decouples LLM knowledge through orthogonal decomposition. It starts by using a probe to analyze how different layers of LLMs are related to specific tasks and how features diverge under varying volatility conditions. Then, it introduces OFDD, which transfers LLM knowledge by breaking it down through layers, tasks, and input data types. Finally, it designs Hájek-MoE to dynamically integrate these decomposed experts, adjusting their contributions based on input data. Extensive experiments demonstrate that CMM outperforms other methods.    

\clearpage

\bigskip

\bibliography{aaai2026}

\end{document}